%% file: acl_latex.tex
\documentclass[11pt]{article}

\usepackage[preprint]{acl}

\usepackage{times}
\usepackage{latexsym}

\usepackage[T1]{fontenc}

\usepackage[utf8]{inputenc}

\usepackage{microtype}
\usepackage{booktabs}
\usepackage{inconsolata}

\usepackage{graphicx}
\usepackage{color}
\usepackage{subcaption}
\usepackage{amsmath}
\usepackage{booktabs}
\usepackage{xcolor}
\usepackage{amssymb}
\usepackage{enumitem} 
\usepackage{multirow}
\usepackage{amsmath, mathrsfs}
\usepackage{bbm}
\usepackage{geometry}
\usepackage{colortbl}
\usepackage{cleveref}
\title{Entropy-Guided Reasoning Compression}

\author{
Hourun Zhu \quad
Yang Gao \thanks{Corresponding author.} \quad
Wenlong Fei \quad
Jiawei Li \quad
Huashan Sun \quad \\
School of Computer Science and Technology \\
Beijing Institute of Technology, Beijing, China \\
\texttt{\{zhuhr, gyang, wenlongfei, jwli, hssun \}@bit.edu.cn}
}

\begin{document}
\maketitle
\begin{abstract}
Large reasoning models have demonstrated remarkable performance on complex reasoning tasks, yet the excessive length of their chain-of-thought outputs remains a major practical bottleneck due to high computation cost and poor deployability. Existing compression methods have achieved partial success but overlook a crucial phenomenon in the training process — the entropy conflict. During compression training, entropy decreases, leading to shorter reasoning but limited exploration, while accuracy-oriented objectives increase entropy, lengthening reasoning chains. This can cause the model to get stuck in a local dilemma. Our analysis further reveals the origin of the entropy conflict: many high-entropy tokens are logical connectors that receive larger gradients and are encouraged under the performance objective, while the compression objective simultaneously penalizes these potentially redundant connectors. This opposing pressure creates a direct source of entropy conflict. To address these issues, we adopt an entropy-guided training framework. As entropy descends, the model is guided toward efficient reasoning by encouraging concise thought steps; as entropy rises, exploration is reinforced under the compact reasoning mode to improve robustness. Experiments on six mathematical benchmarks show that our method compresses reasoning length to 20\% of the original while maintaining or even surpassing baseline accuracy. Code and models will be released publicly.
\end{abstract}

\input{latex/section/Introduction/introduction}
\input{latex/section/Related_work/related_work}

\input{latex/section/Method/method}

\input{latex/section/Experiment/experiment}
\input{latex/section/Conclusion/conclusion}

\bibliography{custom}

\appendix

\input{latex/section/Appendix/appendix}

\end{document}

%% file: latex/section/Introduction/introduction.tex
\section{Introduction}

Large reasoning models (LRMs) have recently shown impressive abilities in mathematical reasoning, scientific question answering, and multi-step decision-making. However, these models often produce unnecessarily long chains of thought. This increases computation cost and reduces reasoning efficiency. Despite several attempts to compress reasoning chains under the RLVR framework, maintaining a proper balance between reasoning efficiency and accuracy remains difficult.

\begin{figure}[t]
  \centering 
  \includegraphics[width=\linewidth]{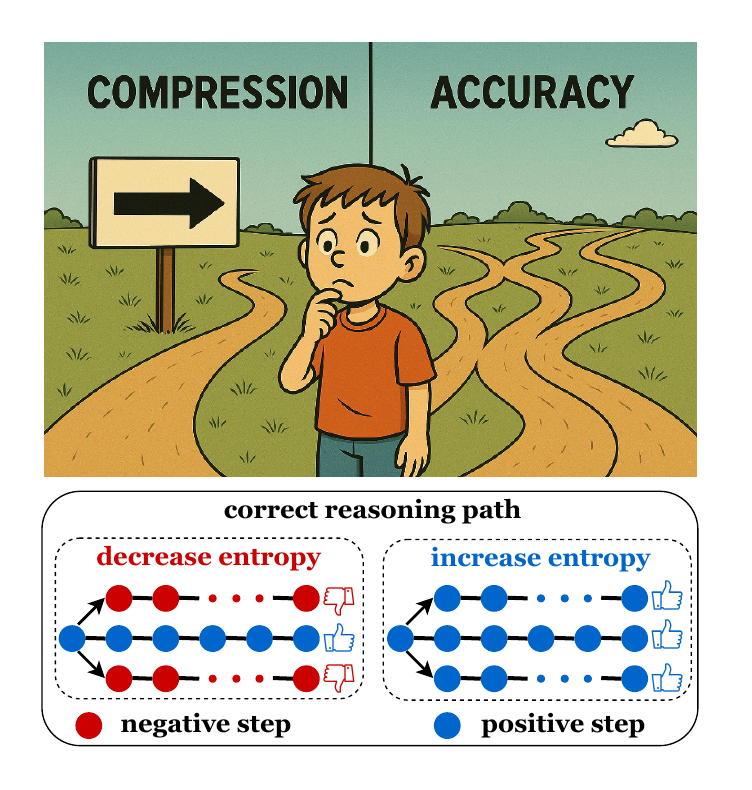}
  \vspace{-2em}
  \caption{Entropy conflict in reasoning training. Compression objectives prefer shorter correct reasoning paths and lower entropy, whereas accuracy objectives accept all correct paths and raise entropy. This creates a fundamental tension between efficiency and correctness.}
  \vspace{-1em}
  \label{fig:entropy_conflict_show}
\end{figure}

In this paper, we identify a key phenomenon that occurs during reasoning-chain compression training, which we call the entropy conflict~\cite{2080rule,hao2025rethinkingentropyinterventionsrlvr}. As shown in Figure~\ref{fig:entropy_conflict_show}, when an LRM is optimized under RLVR-like objectives to shorten its reasoning process, two opposite forces appear. The first force, compression, pushes the model to remove redundant reasoning steps and assign higher probability to concise and efficient reasoning paths. This process reduces entropy in the output distribution. The second force, exploration, encourages the model to search for more diverse reasoning trajectories to improve accuracy. This process increases entropy. These two objectives conflict with each other. Lower entropy improves efficiency but reduces exploration. Higher entropy encourages exploration but causes longer reasoning chains.  We observe that this conflict degrades compression-oriented training: after the model reaches a moderate compression level, further shortening becomes increasingly difficult and quickly stalls.

Our analysis reveals another mechanism that further explains this stagnation. 
During the exploration phase, the gradient term 
\(\nabla_\theta \log \pi_\theta(y)\)
amplifies updates on high-entropy tokens. 
These tokens receive larger gradient magnitudes compared to others, which increases their probability during training. 
By analyzing the token statistics, we find many of these high-entropy tokens are reasoning connectors such as \textit{therefore}, \textit{thus}, and \textit{as a result}. 
As a consequence, these connectors appear more frequently and further lengthen the reasoning chain.
We call this effect reasoning-chain self-extension. 
It drives the model to generate longer and more redundant reasoning chains, which raises entropy again and leads to inefficient reasoning behavior.

Guided by the entropy trajectory observed during training, we structure the learning process into an entropy-descending compression phase and an entropy-ascending exploration phase. In the compression phase, the model is optimized at the sequence level to shorten its reasoning process and internalize a concise response pattern. In the subsequent exploration phase, we increase the sampling temperature to encourage diverse reasoning while applying an accuracy-oriented objective that refines the model’s short-form reasoning. This entropy-guided design naturally decouples compression and exploration, enabling the model to achieve both brevity and strong reasoning performance.

Across multiple mathematical reasoning benchmarks, our method reduces response length by nearly 80\%, while accuracy not only remains preserved but consistently improves. To understand the underlying mechanism, we first analyze the phenomenon of entropy conflict and show that it hinders effective compression by trapping the model in a local optimization regime where further shortening becomes difficult. A token-level diagnostic study further reveals the sources of this conflict. We then conduct detailed ablation experiments to validate the contribution of each component in our method. Finally, case studies demonstrate that our approach successfully suppresses redundant reasoning steps and produces concise yet reliable reasoning traces.

%% file: latex/section/Related_work/related_work.tex

\section{Related work}

\subsection{Large Reasoning Model}

CoT series techniques~\cite{cotelicits,tot,got} endow LLMs with robust reasoning capabilities, enabling them to effectively tackle complex tasks.
Recent state-of-the-art models \cite{openai2024openaio1card,deepseekr1,qwen3,qwq32b,openai2024gpt4ocard}, leveraging their excellent reasoning capabilities, can generate high-quality responses. Nevertheless, in the process of producing such high-quality outputs, these models typically undergo an excessively lengthy reasoning process: they often need to go through multiple rounds of self-doubt and correction before delivering the final answer, and most of the reasoning content during this process is redundant.

\subsection{SFT With Short-Length Dataset}

Some studies introduce external knowledge through fine-tuning to help models learn concise and efficient reasoning strategies, enabling them to balance performance and efficiency. 
For example, Paper \cite{distilling2to1} enables smaller models to imitate the compact reasoning style of large reasoning models. 
Thinkless \cite{thinkless} incorporates external supervision and introduces dual reasoning modes—short mode and thinking mode—to control the depth of reasoning dynamically. 
VeriThinker \cite{verithinker} finetunes models to develop a self-evaluation mechanism that decides whether to stop or continue reasoning. 
Similarly, CoT-Valve \cite{cotvalve} designs a mixed CoT dataset to train models to adaptively adjust their output length based on task difficulty. 
However, these fine-tuning approaches that rely heavily on external datasets often lead to catastrophic forgetting, impairing the model’s inherent reasoning capability.

\begin{figure*}[ht]
    \centering
    \includegraphics[width=\textwidth]{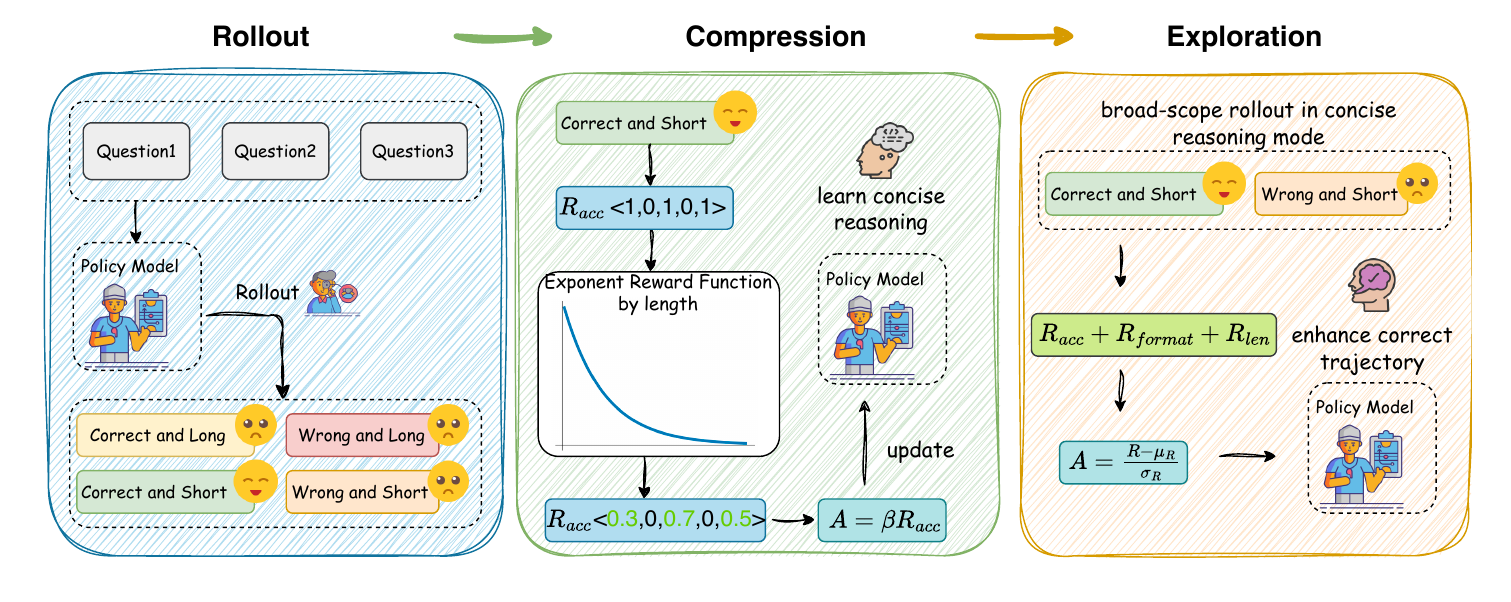}
    \caption{Overview of our method. The model first learns concise reasoning through compression. Then it enhances reasoning ability in the exploration phase via broader rollouts.}
    \label{fig:compress}
\end{figure*}

\subsection{Length Penalty Design}

A straightforward approach to reducing response length is to incorporate a length penalty term into the reward function. 
Several studies~\cite{DAST,L1,demystifying,o1pruner,traininglanguagemodelsreason} successfully compress the model's reasoning process using such penalties, but often at the cost of reduced performance. 
Other methods~\cite{bingo,grpolambda,concisereasoningreinforcementlearning} group samples to jointly optimize for two objectives: shortening the reasoning chain while maintaining or improving model performance. 
This approach streamlines inference and stabilizes performance to some extent. 
From the perspective of entropy dynamics, however, these two objectives correspond to opposing optimization directions. 
Compressing the chain of thought inherently reduces entropy, while enhancing model performance through exploration increases entropy. 
The conflict between these two forces disrupts smooth parameter updates during training, resulting in instability. 
A natural solution is to decouple these objectives into separate stages. 
By optimizing compression and performance independently, this strategy removes inter-task constraints and substantially improves training stability.

%% file: latex/section/Method/method.tex
\definecolor{mypurple}{rgb}{0.949, 0.902, 0.984}
\definecolor{mygreen}{rgb}{0.902, 0.961, 0.902}

\section{Method}

\subsection{Preliminary}

To establish a unified analytical framework, we adopt a general mathematical formulation following \cite{deepseekmath}:

\begin{equation}
    \label{deepseekmath}
    \begin{split}
    &\nabla_{\theta} \mathcal{J}_{\mathcal{A}}(\theta) = \mathbb{E}_{\underbrace{(q, o) \sim \mathcal{D}}_{\text {Data Source }}} \bigg( \frac{1}{|o|} \sum_{t=1}^{|o|} \\
    &\quad \underbrace{G C_{\mathcal{A}}\left(q, o, t, \pi_{r f}\right)}_{\text {Gradient Coefficient }} \nabla_{\theta} \log \pi_{\theta}\left(o_{t} \mid q, o_{<t}\right) \bigg)
    \end{split}
    \nonumber
\end{equation}

In policy optimization, the gradient coefficient is typically defined as the advantage function. 
This means that the model optimizes its parameters by increasing the probability of trajectories with positive advantage 
and decreasing those with negative advantage, thereby maximizing the overall objective. 
When applied to reasoning-chain compression, this process essentially guides the model to identify and reinforce 
a subset of concise and efficient reasoning paths within its original output space.
In the following section, we provide a theoretical analysis of the entropy conflict that arises in this compression process.

Let \( Y \) denote the random variable representing the model output, 
\( C_{acc} \) represent the accuracy-based training condition, 
and \( C_{comp} \) denote an additional compression constraint. 
When the model is trained and converged under the accuracy objective alone, 
its output uncertainty is given by the conditional entropy \( H(Y \mid C_{acc}) \). 
After introducing the compression constraint and retraining to convergence, 
the corresponding entropy becomes \( H(Y \mid C_{acc}, C_{comp}) \).

According to the definition of conditional mutual information, we have:

\begin{equation}
\begin{aligned}
&I(Y; C_{acc} \mid C_{comp}) \\
&= H(Y \mid C_{acc}) - H(Y \mid C_{acc}, C_{comp}).
\end{aligned}
\nonumber
\end{equation}

Because mutual information is always non-negative, so we have
\begin{equation}
H(Y \mid C_{acc}, C_{comp}) \le H(Y \mid C_{acc}).
\nonumber
\end{equation}

With the introduction of a compression constraint, it indicates that compression narrows the output space and reduces entropy.
Conversely, when accuracy-oriented rewards dominate, the model tends to explore longer reasoning paths to improve accuracy, which leads to an increase in entropy.
This reveals an intrinsic opposition between accuracy optimization and reasoning compression—an entropy conflict that underlies instability in joint training. Further discussion can be found in \ref{apx:discussion}.

\subsection{Overview}

Guided by the entropy dynamics observed in reasoning models, our method operates through an entropy-descending compression phase followed by an entropy-ascending exploration phase.
In the compression phase, the model is optimized to suppress redundant reasoning by directly regularizing response length. Because reasoning length closely correlates with the number of reasoning steps, overly long or incorrect responses are discarded, while correct and concise ones are reinforced. The advantage of each positive sample is exponentially decayed with respect to its length, assigning larger gradients to shorter and more efficient reasoning paths. This process concentrates probability mass on concise reasoning trajectories and substantially lowers model entropy.

In the subsequent exploration phase, we reuse GRPO with a higher sampling temperature and relaxed exploration constraints to restore reasoning diversity. Accuracy-oriented rewards dominate this phase, while only severe redundancy is mildly penalized, enabling the model to refine short-form reasoning without collapsing into overly deterministic patterns.

\subsection{Compressing Stage}

The first step of our framework is to teach LRMs to think in a concise and efficient manner. 

\subsubsection{Length Clipping}

To encourage the model to produce concise reasoning trajectories, we introduce a Length Clipping mechanism during sampling. When a generated reasoning sequence exceeds a predefined maximum length 
$L$ without producing a final answer token, the trajectory is terminated and assigned a zero reward. Formally, the clipped reward is defined as:
\begin{equation}
R_{\text{clip}} =
\begin{cases}
0, & \text{if } |y| > L , \\
R(y), & \text{otherwise.}
\end{cases}
\label{eq:length_clip}
\nonumber
\end{equation}

Since response length is roughly proportional to the number of reasoning steps, overlong outputs often reflect redundant reasoning. By removing trajectories that exceed the length threshold, we filter out samples with a low density of useful reasoning. This clipping mechanism increases the proportion of concise, answer-producing trajectories in the sampled set, allowing subsequent policy updates—especially under positive-only optimization—to rely on higher-quality reasoning examples.

\subsubsection{Positive-Only and Absolute-Advantage Updates}
Previous study \cite{deng2025effectnegativegradientgroup} has reported an interesting phenomenon. Some negative samples are highly similar to positive ones. As a result, their gradients may only slightly increase the probability of positive samples. In some cases, they can even decrease it.
Therefore, to ensure training efficiency, we update the model using only the gradients from the positive samples of the target dataset. The training objective can be expressed as 

\begin{equation}
\label{eq:correct_output}
\nabla_\theta J(\theta) = \mathbb{E}_{y \sim \pi_\theta} \left[ \nabla_\theta \log \pi_\theta(y) \cdot A(y) \right]_{R(y) > 0}.
\nonumber
\end{equation}

the advantage function is computed as $A(y) = \beta R(y)$,
where $\beta$ denotes the scaling coefficient. We adopt this absolute advantage formulation instead of the relative advantage used in standard GRPO. 
This design choice helps to prevent instability introduced by normalization.
In particular, when relative normalization is applied, certain correct trajectories with rewards slightly below the batch mean may be incorrectly treated as negative examples, thus introducing misleading penalty gradients. 
By contrast, the absolute advantage preserves the positivity of genuinely correct samples and yields more stable and reliable policy updates.

\begin{table*}[th!]
\centering
\resizebox{\textwidth}{!}{
\begin{tabular}{l | ccc | ccc | ccc | ccc | ccc | ccc}
\toprule
\textbf{Model Name} & \multicolumn{3}{c|}{\textbf{AIME24}} & \multicolumn{3}{c|}{\textbf{Minerva}} & \multicolumn{3}{c|}{\textbf{MATH}} & \multicolumn{3}{c|}{\textbf{GSM8K}} & \multicolumn{3}{c|}{\textbf{AMC}} & \multicolumn{3}{c}{\textbf{OlymBench}} \\
\cmidrule(lr){2-19}
 & ACC & LEN & LAC & ACC & LEN & LAC & ACC & LEN & LAC & ACC & LEN & LAC & ACC & LEN & LAC & ACC & LEN & LAC \\
\midrule
Qwen2.5-1.5B-Ins & 0.0 & 1300 & - & 77.7 & 993 & - & 51.7 & 855 & - & 70.2 & 466 & - & 0 & 730 & - & 7.3 & 682 & - \\
Qwen2.5-1.5B-Math-Ins & 11.3 & 1128 & - & 91.8 & 586 & - & 76.0 & 721 &  -& 85.7 & 447 & - & 0 & 823 & - & 21.4 & 774 & - \\
\midrule
R1-Distill-Qwen2.5-1.5B & 33.3 & 16927 & -- & 96.4 & 3054 & -- & 87.2 & 5485 & -- & 84.6 & 1909 & -- & 61.4 & 11157 & -- & 44.7 & 11236 & -- \\
\rowcolor[rgb]{0.949, 0.902, 0.984}AutoThink & 30.0 & 7016 & 23.0 & 95.5 & 959 & 79.1 & 87.0 & 1851 & 70.8 & 86.4 & 743 & 67.5 & \textbf{66.3} & 4656 & 50.6 & 45.2 & 5296 & 32.8 \\
\rowcolor[rgb]{0.949, 0.902, 0.984} AdaptThink &33.3 &8492 &23.5 &96.3 &1267 &73.7 &82 &2158 &63.9 &83.1 &971 &58.3 &62.6 &4194 &49.5 &44.5 &5291 &32.4 \\

\rowcolor[rgb]{0.949, 0.902, 0.984}Thinkless & 36.7 & 8848 & 25.3 & 95.3 & 1178 & 74.7 & 84.6 & 2623 & 61.1 & 84.5 & 637 & 69.0 & 57.8 & 6861 & 35.9 & 42.5 & 6625 & 27.2 \\
\rowcolor[rgb]{0.902, 0.961, 0.902}ThinkPrune & 26.7 & 8175 & 19.2 & 96.0 & 1403 & 70.6 & 85.8 & 2321 & 65.2 & 85.0 & 907 & 61.6 & 63.9 & 4354 & 49.9 & 43.0 & 4401 & 33.5 \\
\rowcolor[rgb]{0.902, 0.961, 0.902}Laser & \textbf{40.0} & 5635 & 32.7 & 96.6 & 1323 & 72.7 & 87.0 & 2073 & 68.6 & 86.2 & 996 & 59.6 & 65.1 & 4963 & 48.5 & \textbf{45.8} & 5512 & 32.7 \\
\rowcolor[rgb]{0.902, 0.961, 0.902}LC-R1 & 23.3 & 11124 & 13.7 & 94.4 & 1374 & 70.0 & 83.6 & 2921 & 57.2 & 80.1 & 789 & 61.3 & 60.2 & 4778 & 45.5 & 39.3 & 4639 & 30.1 \\
\rowcolor[rgb]{0.902, 0.961, 0.902}ours & 36.7 & \textbf{3634} & \textbf{32.5} & \textbf{97.1} & \textbf{671} & \textbf{85.7} & \textbf{87.4} & \textbf{1180} & \textbf{77.4} & \textbf{87.2} & \textbf{586} & \textbf{72.6} & \textbf{66.3} & \textbf{2083} & \textbf{59.8} & 45.3 & \textbf{2275} & \textbf{40.5} \\

\midrule
Qwen2.5-7B-Ins & 12.0 & 1016 & -  & 91.2 & 649 & - & 74.2 & 567 & - & 90.9 & 279 & - & 47.5 & 801 & - & 39.2 & 827 & -  \\
Qwen2.5-7B-Math & 19.0 & 1429 & -  & 93.1 & 761 & - & 63.4 & 740 & -& 93.2 & 439 & -   & 62.5 & 1022 & - & 31.5 & 1037 & -  \\
Qwen2.5-7B-Math-Ins & 10.3 & 1363 & -  & 94.2 & 730 & - & 81.4 & 670 & - & 95.2 & 323 & - & 60.0 & 1029 & - & 38.9 & 1027 & - \\

\midrule

R1-Distill-Qwen2.5-7B & 56.7 & 14108 & - & 98.7 & 2435 & - & 93.4 & 4181 & - & 92.8 & 1700 & - & 81.9 & 7630 & - & 56.2 & 9031 & - \\
\rowcolor[rgb]{0.949, 0.902, 0.984}AutoThink & \textbf{56.7} & 6611 & 41.3 & \textbf{98.6} & 944 & 77.2 & 92.6 & 1808 & 69.8 & 93.4 & 695 & 71.8 & \textbf{84.3} & 3638 & 60.7 & \textbf{55.0} & 4415 & 42.9 \\
\rowcolor[rgb]{0.949, 0.902, 0.984}AdaptThink &53.3 &8023 &35.0 &98.1 &1139 &71.6 &92 &1902 &67.9 &91 &812 &65.8 &83.1 &3891 &58.2 &52.8 &3782 &43.0 \\
\rowcolor[rgb]{0.902, 0.961, 0.902}Laser & 50.0 & 6054 & 37.8 & 98.3 & 1276 & 67.8 & \textbf{93.0} & 1762 & 70.7 & 93.0 & 965 & 61.2 & 81.9 & 3437 & 60.7 & 54.3 & 3365 & 45.5 \\
\rowcolor[rgb]{0.902, 0.961, 0.902}LC-R1 & 46.7 & 7384 & 32.2 & 97.1 & 850 & 78.4 & 92.4 & 1653 & 71.8 & 90.2 & 507 & 75.6& 80.7 & 3418 & 60.0 & 53.0 & 4093 & 42.3 \\
\rowcolor[rgb]{0.902, 0.961, 0.902}ours & 53.3 & \textbf{4094} & \textbf{44.9} & 98.0 & \textbf{651} & \textbf{83.9} & 92.4 & \textbf{1078} & \textbf{79.6} & \textbf{93.6} & \textbf{428} & \textbf{81.0} & 83.1 & \textbf{1838} & \textbf{72.4} & 53.5 & \textbf{1911} & \textbf{48.6} \\

\bottomrule
\end{tabular}
}
\caption{Performance comparison across mathematical reasoning benchmarks. In the table, ``\raisebox{0.5pt}{\colorbox{mypurple}{\phantom{\rule{5pt}{5pt}}}}'': methods directly compressing reasoning chains,
``\raisebox{0.5pt}{\colorbox{mygreen}{\phantom{\rule{5pt}{5pt}}}}'': methods learning adaptive reasoning behaviors, \textbf{Bold}: best performance among non-base models. }
\label{tab:main_results}

\end{table*}

\subsubsection{Exponent Reward Shaping}

Reward shaping is often one of the most crucial components in the design of reinforcement learning algorithms. With properly designed reward shaping, the model can continuously refine and adapt its generation style.
To enable the model to efficiently acquire strong reasoning ability, we employ an exponential function as the reward shaping function as Equation \ref{eq:exponent}.
This function is designed to allocate larger gradients to relatively shorter responses while reducing the gradient contribution of longer ones. The degree of compression can be controlled by the parameters $L$ and $r$, where $L$ denotes the response length and $r$ specifies the reward value when the length equals $L$.

\begin{equation}
\label{eq:exponent}
\begin{aligned}
f(x) &= e^{-\lambda x}, \\
\lambda &= -\frac{\ln r}{L}.
\end{aligned}
\nonumber
\end{equation}

This formulation amplifies gradients for shorter samples while compressing them for longer ones, naturally guiding the model to prioritize shorter and more efficient reasoning patterns.

\subsection{Enhancing Stage}

After the model has learned concise reasoning, we increase the sampling temperature and broaden the sampling range to encourage freer exploration, thereby further improving its reasoning accuracy and format robustness.

In this stage, the policy is optimized over all sampled trajectories using the conventional GRPO loss:

\begin{equation}
\mathcal{L}_{\text{GRPO}}(\theta)
= - \mathbb{E}_{y \sim \pi_\theta} 
\left[
\frac{\pi_\theta(y)}{\pi_{\theta_{\text{old}}}(y)} 
\cdot A(y)
\right],
\label{eq:grpo_loss}
\nonumber
\end{equation}

where $\pi_\theta$ and $\pi_{\theta_{\text{old}}}$ denote the current and reference policies, respectively. 
The advantage function is computed using the relative advantage formulation:

\begin{equation}
A(y) = \frac{R(y) - \mu_R}{\sigma_R},
\label{eq:relative_advantage}
\nonumber
\end{equation}

where $\mu_R$ and $\sigma_R$ represent the batch mean and standard deviation of the rewards, respectively.

To provide richer and more interpretable supervision signals, we decompose the total reward into multiple components:
\begin{equation}
R(y) = R_{\text{format}} 
+ R_{\text{answer}}
+ R_{\text{length}}.
\nonumber
\end{equation}

To discourage excessively long reasoning, we set $R_length$ to decay by 0.25 for every additional 1k tokens beyond the length clipping threshold.


%% file: latex/section/Experiment/experiment.tex
\section{Experiment}

\begin{figure*}[!t]
    \centering

    \begin{subfigure}[t]{0.48\textwidth}
        \centering
        \includegraphics[width=\linewidth]{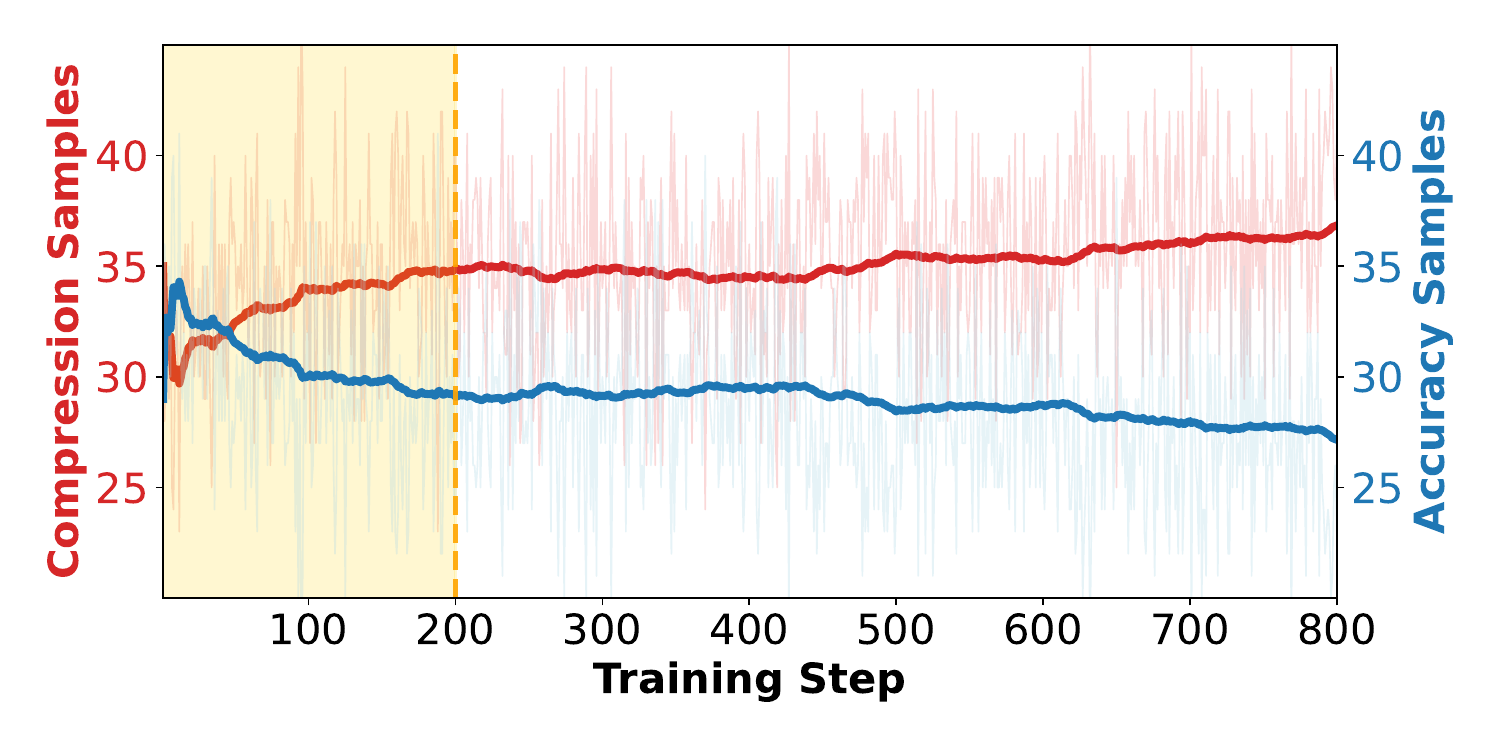}
        \caption{}
        \label{fig:acc_cmp}
    \end{subfigure}
    \hfill
    \begin{subfigure}[t]{0.48\textwidth}
        \centering
        \includegraphics[width=\linewidth]{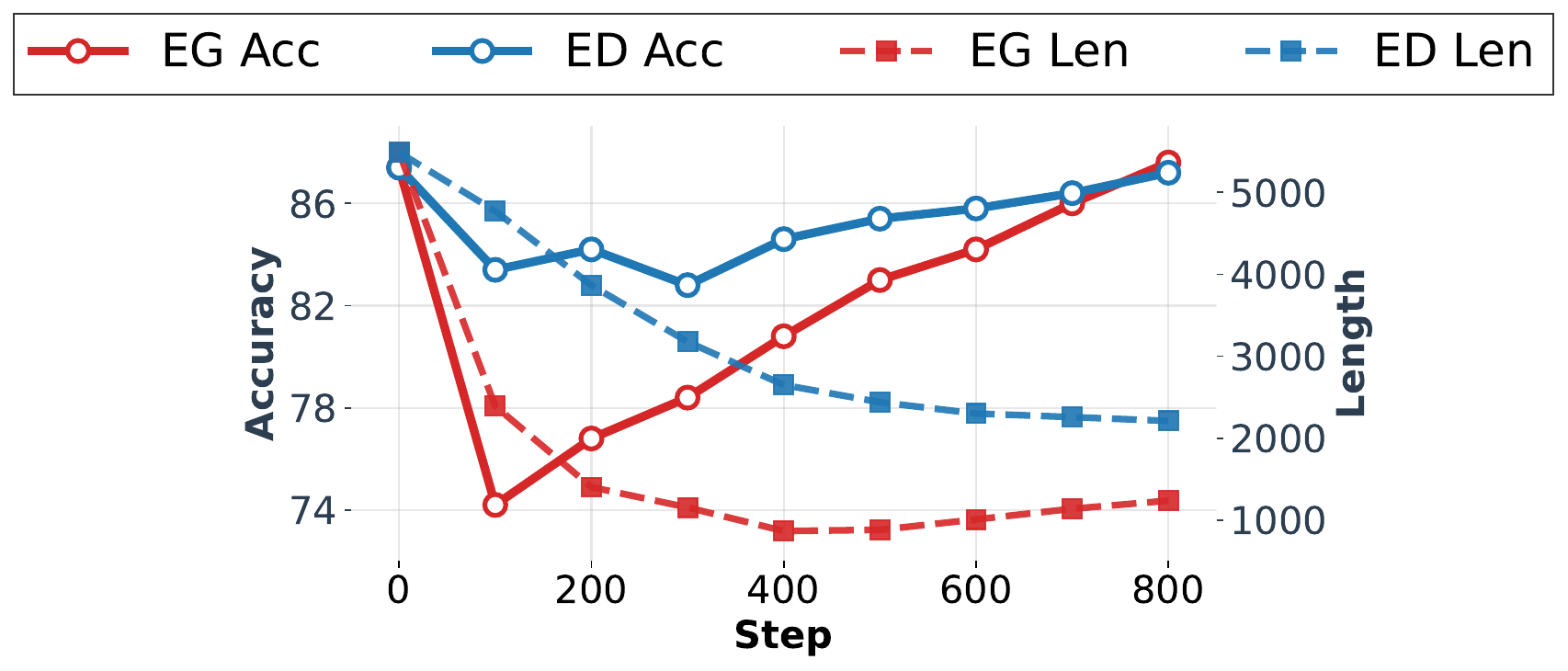}
        \caption{}
        \label{fig:entropy_conflict_performance}
    \end{subfigure}


    \begin{subfigure}[b]{0.47\textwidth}
        \centering
        \resizebox{\linewidth}{!}{ 
        \begin{tabular}{l cc cc cc cc}
            \toprule
            \multirow{2}{*}{Model Name}
                & \multicolumn{2}{c}{AIME24}
                & \multicolumn{2}{c}{Minerva}
                & \multicolumn{2}{c}{MATH}
                & \multicolumn{2}{c}{GSM8K} \\
            \cmidrule(lr){2-3}
            \cmidrule(lr){4-5}
            \cmidrule(lr){6-7}
            \cmidrule(lr){8-9}
            & ACC & LEN & ACC & LEN & ACC & LEN & ACC & LEN \\
            \midrule
            Base & 33.3 & 16927 & 96.4 & 3054 & 87.2 & 5485 & 84.6 & 1909 \\
            EG  & 36.7 & 3634 & 97.1 & 671  & 87.4 & 1180 & 87.2 & 586 \\
            EE  & 33.3 & 7134 & 96.6 & 1168 & 87.2 & 2231 & 85.4 & 871 \\
            EE* & 40.0 & 6721 & 97.3 & 982  & 87.6 & 1992 & 86.8 & 714 \\
            \bottomrule
        \end{tabular}}
        \vspace{1em}
        \caption{}
        \label{tab:entropy_conflict_performance}
    \end{subfigure}
    \hfill
    \begin{subfigure}[b]{0.24\textwidth}
        \centering
        \includegraphics[width=\linewidth]{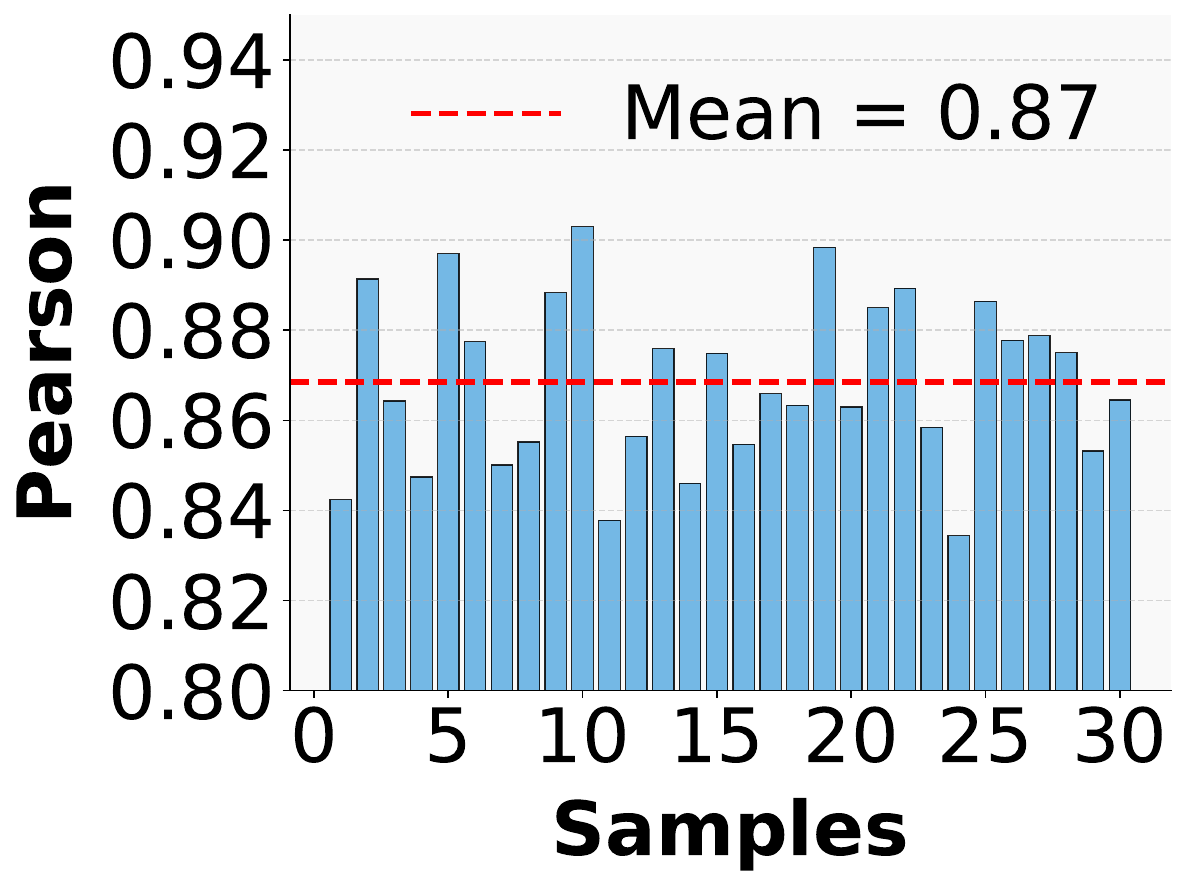}
        \caption{}
        \label{fig:pearson}
    \end{subfigure}
    \begin{subfigure}[b]{0.24\textwidth}
        \centering
        \includegraphics[width=\linewidth]{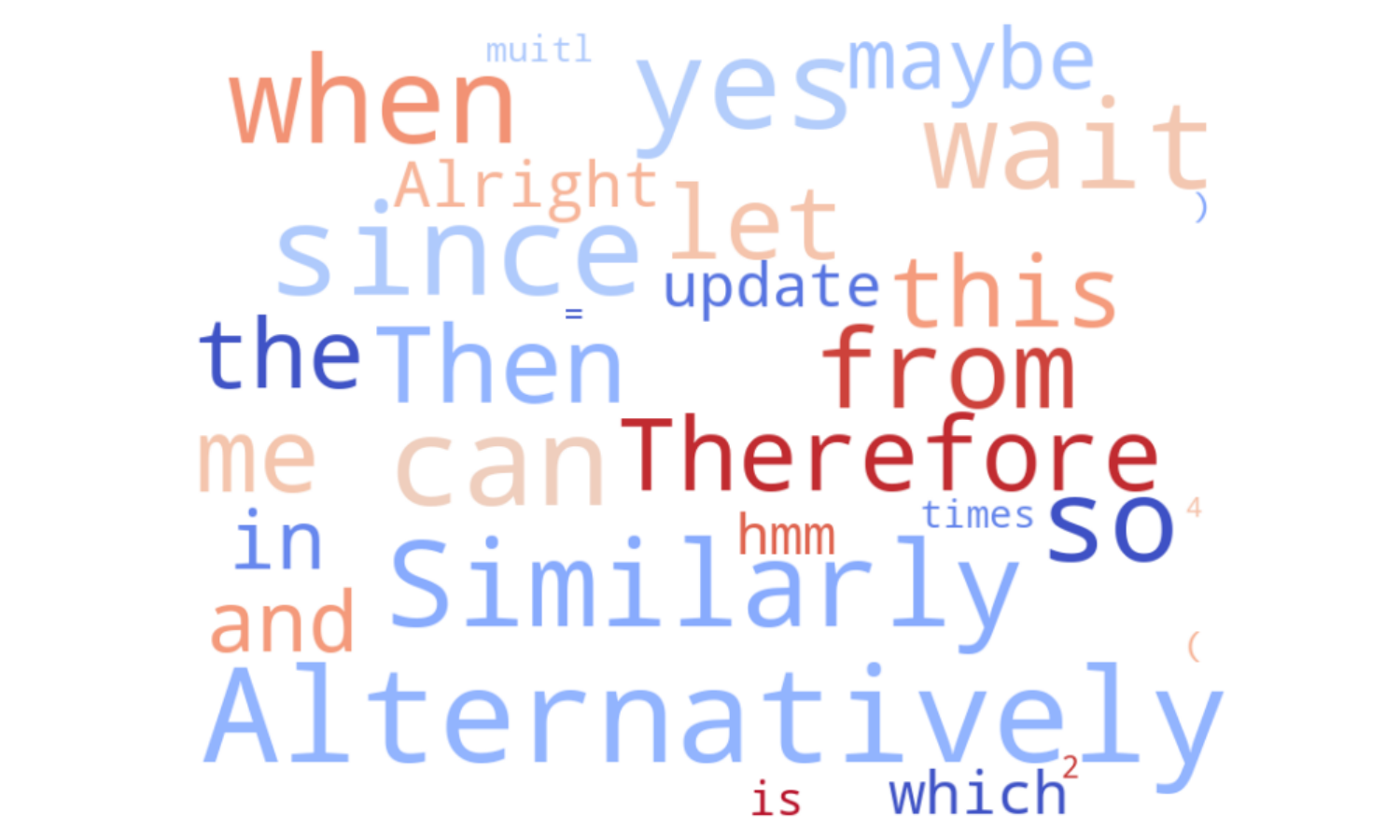}
        \caption{}
        \label{fig:word}
    \end{subfigure}

    \caption{Overall empirical evidence of entropy conflict.
(a) Distribution of samples selected by the compression objective (red) and accuracy objective (blue) during training. The detailed setup is provided in Section \ref{entropy_conflict_setup}.
(b) Performance comparison on MATH under Entropy-Guided (EG) and Entropy-Entangled (EE) training. 
(c) Evaluation of EG, EE, and an extended EE* variant (trained for 400 additional steps) across four benchmarks (AIME24, Minerva, MATH, GSM8K).
(d) Pearson correlation between token-level entropy and gradient magnitude during training.
(e) Word-cloud visualization of some high-entropy tokens frequently encountered during training.}
    \label{fig:bigblock}
\end{figure*}

\subsection{Experimental Setups}

We use DeepSeek-R1-Distill-Qwen-1.5B and DeepSeek-R1-Distill-Qwen-7B as the base model and perform reinforcement learning training on the DeepScaleR-Preview \cite{deepscaler2025} dataset.
The detailed training configurations for both models are provided in the appendix~\ref{apx:impl}.
We evaluate the trained models on six mathematical benchmarks, measuring both Pass@1 accuracy and response length. The benchmarks include: AIME24 \cite{aime24}, AMC, MATH \cite{math500}, MinervaMATH \cite{Minerva}, Olympiad-Bench \cite{olympiadbench}, GSM-8K \cite{gsm8k}. To better assess the model’s reasoning efficiency, we introduce a key metric for evaluation from \cite{bingo}.

\begin{equation}
\mathrm{L\text{-}Acc} = \mathrm{Acc} \times \sqrt{1 - \frac{L}{L_{base}}} \quad,
\nonumber
\end{equation}

where $L$ is the average response length and $L_{base}$ is the length of base model.

The baselines we selected are currently open-source methods with strong performance, namely ThinkPrune \cite{thinkprune}, Laser \cite{laser}, AutoThink \cite{autothink}, LC-R1 \cite{lcr1}, and Thinkless \cite{thinkless}. When calculating response length, we followed the approach in Thinkless and modified the chat templates for all models, including the content within the <think> tags in the length measurement.

\subsection{Main Results}

\textbf{Reasoning Significantly Improves Mathematical Problem-Solving.}
Across both 1.5B and 7B scales, chain-of-thought reasoning improves mathematical accuracy.
Reasoning-enabled models solve problems step by step.
This structured process helps them decompose complex problems and reach more reliable solutions.

\textbf{Adaptive Reasoning shows limited compression, Direct Compression suffers performance loss.}
Different compression strategies exhibit distinct trade-offs. Adaptive reasoning models maintain stable accuracy but achieve limited reduction in reasoning length, indicating a weaker compression effect. In contrast, directly shortening reasoning chains yields stronger length compression but often leads to noticeable performance degradation. This contrast highlights the underlying entropy conflict between accuracy and compression objectives, emphasizing the challenge of balancing efficiency and reasoning quality.

\textbf{Entropy-Guided Compression achieves optimal balance between response length and reasoning accuracy.}
Our proposed entropy-guided reasoning compression method achieves optimal results across benchmarks. It effectively reduces reasoning length while preserving high accuracy, showing that the model attains a dynamic balance between compression and expressiveness, validating the effectiveness of our entropy conflict resolution strategy.

\subsection{Empirical Analysis of Entropy Conflict in Reasoning Compression}

\label{entropy_conflict_setup}

To examine whether entropy conflict emerges in practice, we construct an entropy-coupled setup. In each batch, samples are split into two groups based on their sampled accuracy. High-accuracy($\geq 50\%$) samples are treated as compression samples and low-accuracy($< 50\%$) samples are used for accuracy optimization.

\begin{table*}[h]
\centering
\resizebox{0.9\linewidth}{!}{
\begin{tabular}{lcccccccc}
\toprule
\multirow{2}{*}{Model} &
\multicolumn{2}{c}{AIME24} &
\multicolumn{2}{c}{Minerva} &
\multicolumn{2}{c}{MATH} &
\multicolumn{2}{c}{GSM8K} \\
\cmidrule(lr){2-3}
\cmidrule(lr){4-5}
\cmidrule(lr){6-7}
\cmidrule(lr){8-9}
& ACC & LEN & ACC & LEN & ACC & LEN & ACC & LEN \\
\midrule
Base & 33.3 & 16927 & 96.4 & 3054 & 87.2 & 5485 & 84.6 & 1909 \\
\midrule
\multicolumn{9}{l}{\textbf{Stage Ordering Ablation}} \\
Exploration → Compression (after exploration)  
& 40.0 & 15127 & 97.5 & 2953 & 88.6 & 5135 & 87.9 & 2202 \\
Exploration → Compression (after compression)  
& 30.0 & 3763 & 94.0 & 714 & 83.6 & 1089 & 80.9 & 460 \\
Compression → Exploration (ours)
& 36.7 & 3634 & 97.1 & 671 & 87.4 & 1180 & 87.2 & 586 \\
\midrule
\multicolumn{9}{l}{\textbf{Length Clip Ablation}} \\
LC = 2048  & 33.3 & 3418 & 94.8 & 590 & 84.6 & 1017 & 85.1 & 412 \\
LC = 4096 (ours) & 36.7 & 3634 & 97.1 & 671 & 87.4 & 1180 & 87.2 & 586 \\
LC = 8192  & 43.3 & 6662 & 97.3 & 829 & 87.6 & 1690 & 87.3 & 700 \\
\midrule
\multicolumn{9}{l}{\textbf{Reward Function Ablation}} \\
cosine  & 36.7 & 5705 & 96.6 & 871 & 86.2 & 1428 & 86.5 & 683 \\
linear  & 33.3 & 4783 & 96.5 & 761 & 86.8 & 1509 & 86.0 & 612 \\
exponent (ours) & 36.7 & 3634 & 97.1 & 671 & 87.4 & 1180 & 87.2 & 586 \\
\bottomrule
\end{tabular}}
\caption{Comprehensive ablation study including stage ordering, length-clip threshold, and reward shaping. All decimals are rounded to one digit.}
\end{table*}

\textbf{Training stagnation reveals the presence of entropy conflict.}
To examine how compression and accuracy objectives interact during training, we track the number of samples selected by each objective (Figure \ref{fig:acc_cmp}).
At the beginning, both curves fluctuate as the model adjusts its reasoning behavior. Within the first 200 steps, compression-selected samples gradually increase, while accuracy-selected samples steadily decrease. The model begins to favor shorter reasoning paths.

After 200 steps, both curves flatten and remain almost unchanged. This stagnation marks the first clear sign of entropy conflict. The compression objective pushes the model toward shorter chains, while the accuracy objective favors longer correct paths that compression penalizes. These two goals pull in opposite directions and trap the optimization in a difficult regime, where further compression becomes hard to achieve.

\textbf{Entropy conflict blocks further compression.}
We evaluate how entropy conflict affects actual model performance using both training curves and benchmark results.
Figure \ref{fig:entropy_conflict_performance} shows that under entropy-entangled training, the model maintains reasonable accuracy but fails to continue compressing after the early stage. The length curve nearly stops decreasing after 400 steps, indicating that compression is fundamentally constrained by the competing objective.

In contrast, the entropy-guided method achieves much stronger compression while recovering accuracy later in training. The separation of objectives prevents the mutual interference that stalls entropy-entangled training.

Figure \ref{tab:entropy_conflict_performance} further confirms this pattern on four benchmarks. Entropy-guided consistently produces shorter reasoning chains and stronger overall performance. The extended entropy-entangled model (EE* in the table) receives 400 additional training steps.
However, its reasoning length decreases only slightly.
This result shows that the entangled objective fundamentally limits the model’s compression capability. These results show that entropy conflict directly harms both compression efficiency and downstream accuracy.

\textbf{Token-Level Mechanism of Entropy Conflict.}
To understand why entropy conflict happens under entropy-entangled training, we analyze token-level behavior during optimization. Figure \ref{fig:pearson} shows a strong correlation between token entropy and gradient magnitude. High-entropy tokens consistently receive larger gradients, which means they are reinforced more strongly during training.
We further inspect the distribution of these tokens  as Figure \ref{fig:word}. Apart from some context-specific words tied to the problem description, many high-entropy tokens are reasoning connectors such as Then, Therefore, Similarly, and Alternatively. These tokens typically appear at the beginning of a reasoning step and often introduce a new segment of the chain.


\begin{figure*}[t]
    \centering
    \begin{subfigure}{0.48\textwidth}
        \centering
        \includegraphics[width=\linewidth]{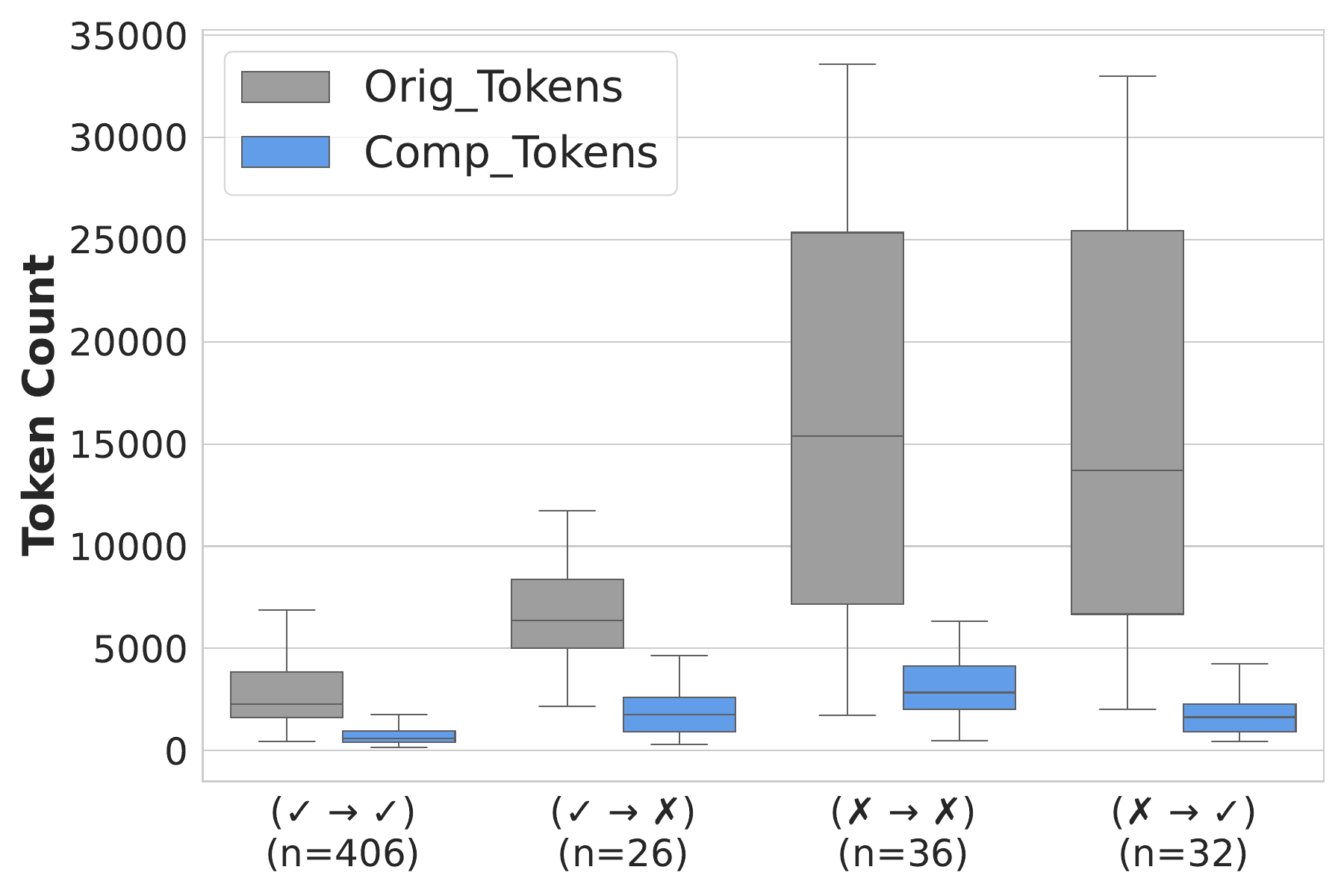}
        \caption{Token Distribution by Group}
        \label{fig:token_dist}
    \end{subfigure}
    \hfill
    \begin{subfigure}{0.48\textwidth}
        \centering
        \includegraphics[width=\linewidth]{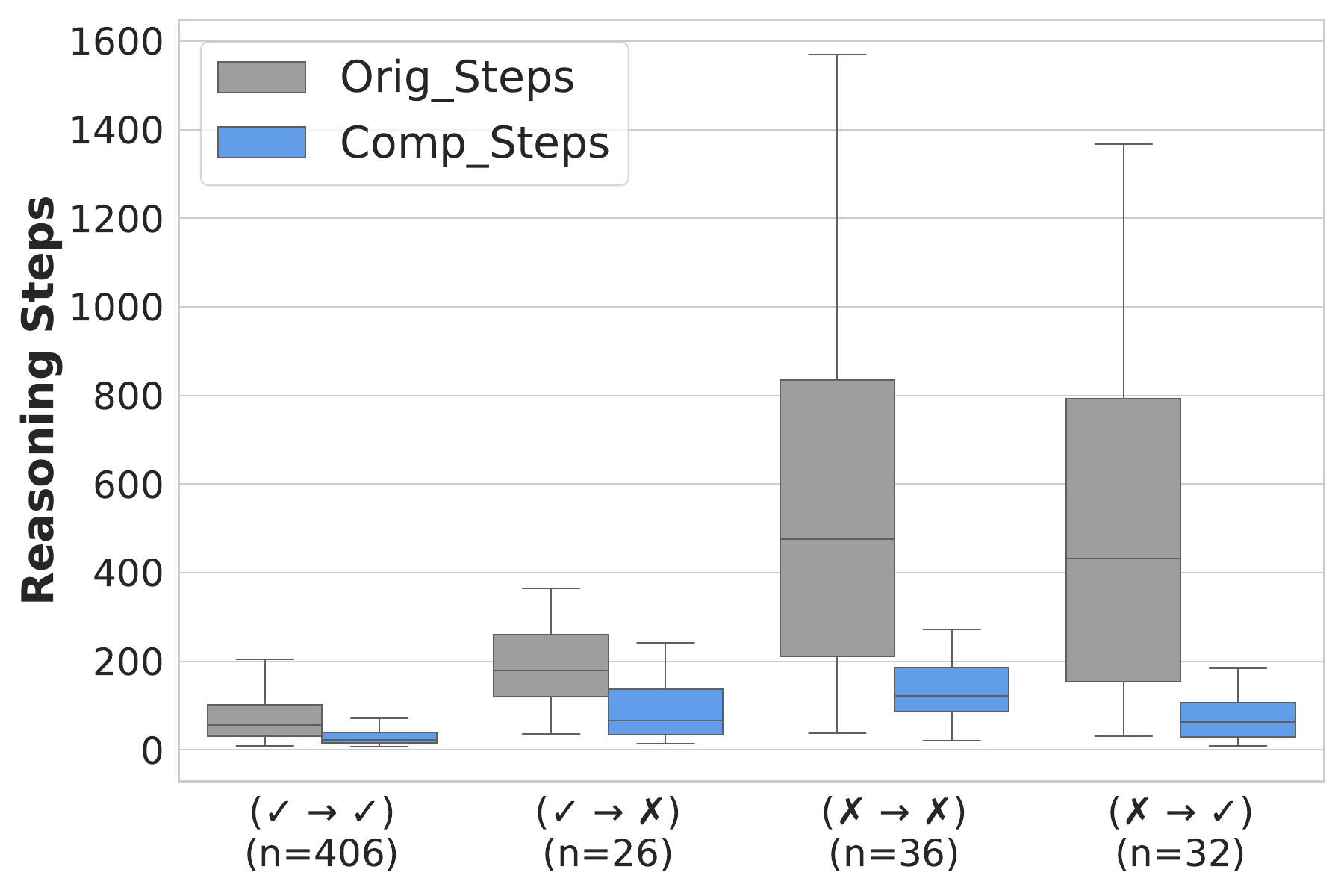}
        \caption{Reasoning Steps Distribution by Group}
        \label{fig:step_dist}
    \end{subfigure}
    \caption{Comparison of the original model and our compressed model on the Math500 benchmark, grouped by four correctness transitions: preserved (\checkmark→\checkmark), lost (\checkmark→\texttimes), gained (\texttimes→\checkmark), and failed (\texttimes→\texttimes). The boxplots show the distribution of (a) token counts and (b) reasoning steps for both models within each group.}
    \label{fig:case}
\end{figure*}

When such connectors receive large gradients, their sampling probability increases. As they become more likely to appear, the model generates more intermediate steps, and the reasoning chain naturally grows longer. Longer chains also contain more uncertain tokens, which raises the overall entropy. During compression, these connectors are considered redundant and are penalized. However, under the accuracy objective, the same trajectories are treated as positive examples, and the connectors are reinforced again. 
This token-level behavior explains why entropy conflict arises: the optimization repeatedly strengthens the very tokens that expand the reasoning chain, while the compression objective attempts to suppress them.

\subsection{Ablation Study}

\textbf{Proper stage ordering is crucial for effective compression.}
We first examine the reversed order, where exploration comes before compression. The exploration stage helps the model adapt to its current reasoning pattern and can temporarily improve accuracy. However, the subsequent compression stage disrupts this adapted pattern. It forces the model to rewrite its reasoning structure, which causes a sharp drop in accuracy. This loss is hard to recover within the compression stage.

In contrast, our method uses the order “compression → exploration.” The model first learns a concise reasoning pattern during compression. The exploration stage then adapts the model within this already compressed regime, rather than within a verbose one. This comparison shows that stage ordering is not symmetric: exploration must operate on a compressed reasoning mode; otherwise, compression will break the adapted behavior and severely hurt accuracy.

\textbf{Length clip controls the trade-off between compression and accuracy.}
The length-clip threshold plays a central role in shaping the model’s reasoning style.
A shorter threshold forces the model to drop redundant reasoning steps and quickly learn concise responses. This leads to strong compression but also reduces accuracy, as useful intermediate steps may be truncated.
A larger threshold preserves more reasoning structure and yields higher accuracy, but the model retains lengthy and sometimes unnecessary chains, limiting compression.
Our choice of 4096 strikes a middle ground: it encourages concise reasoning while maintaining stable performance across benchmarks.

\textbf{Reward Shaping Controls the Aggressiveness of Compression.}
Different reward functions impose different sensitivities to reasoning length.
Cosine assigns larger rewards to mid-length samples, so its compression behavior is conservative.
Linear treats lengths more uniformly and produces moderate compression.
Exponent focuses strongly on short samples and heavily rewards concise reasoning.
As a result, exponent yields the most aggressive compression and achieves the best balance between length reduction and accuracy.

\subsection{Case Study}

Our results on Math500 reveal that the proposed compression method not only shortens the generated reasoning traces in terms of token length, but also effectively removes redundant reasoning steps. As shown in Figure~\ref{fig:case}, the responses of the original model are already fairly concise on problems it answers correctly, indicating that it does not engage in excessive or unnecessary reasoning when it can handle the problem confidently. Our training further compresses these already-short traces while largely preserving correctness. However, this compression can occasionally remove useful intermediate reasoning, leading to a small portion of “\checkmark→\texttimes” cases where correct answers become incorrect.

Interestingly, we also observe the opposite trend: a nontrivial number of samples fall into the “\texttimes→\checkmark” group. After compression, the model becomes more accurate because eliminating noisy or misleading reasoning steps helps the model focus on the essential logical structure of the problem. This phenomenon suggests that redundant reasoning is not merely inefficient—it may actually harm accuracy. Further inspection \ref{apx:case} reveals that the compressed model has acquired a more efficient reasoning strategy. During the internal think phase, it no longer performs explicit calculations; instead, it produces only high-level procedural guidance. The concrete computations are carried out in the final output stage, where the model executes the necessary steps rapidly and with minimal redundancy. This behavioral shift indicates that the model has learned to separate planning from execution, leading to substantially more concise and purpose-driven reasoning traces.

%% file: latex/section/Conclusion/conclusion.tex
\section{Conclusion}

Our work reveals the existence of an entropy conflict in the compression training of reasoning models—a phenomenon that hinders training efficiency and prevents full performance development. By decoupling this process, we provide a detailed analysis of how reasoning chains evolve under compression. This understanding enables efficient reasoning-chain compression even with limited data, demonstrating that controlling entropy dynamics is crucial for achieving both compact and high-performing reasoning models.

%% file: latex/section/Appendix/appendix.tex
\section{Implementation Details}
\label{apx:impl}

We implement our training pipeline based on the Verl framework \cite{verl}. During optimization, we use a batch size of 64 and update the model every 16 samples, with eight sampled responses per input to estimate the policy gradient. We adopt the AdamW optimizer with a learning rate of 1e-6. The entropy-guided training consists of two consecutive stages: a compression stage of 400 steps followed by an exploration stage of 400 steps.

\begin{figure*}[ht]
    \centering
    \includegraphics[width=\textwidth]{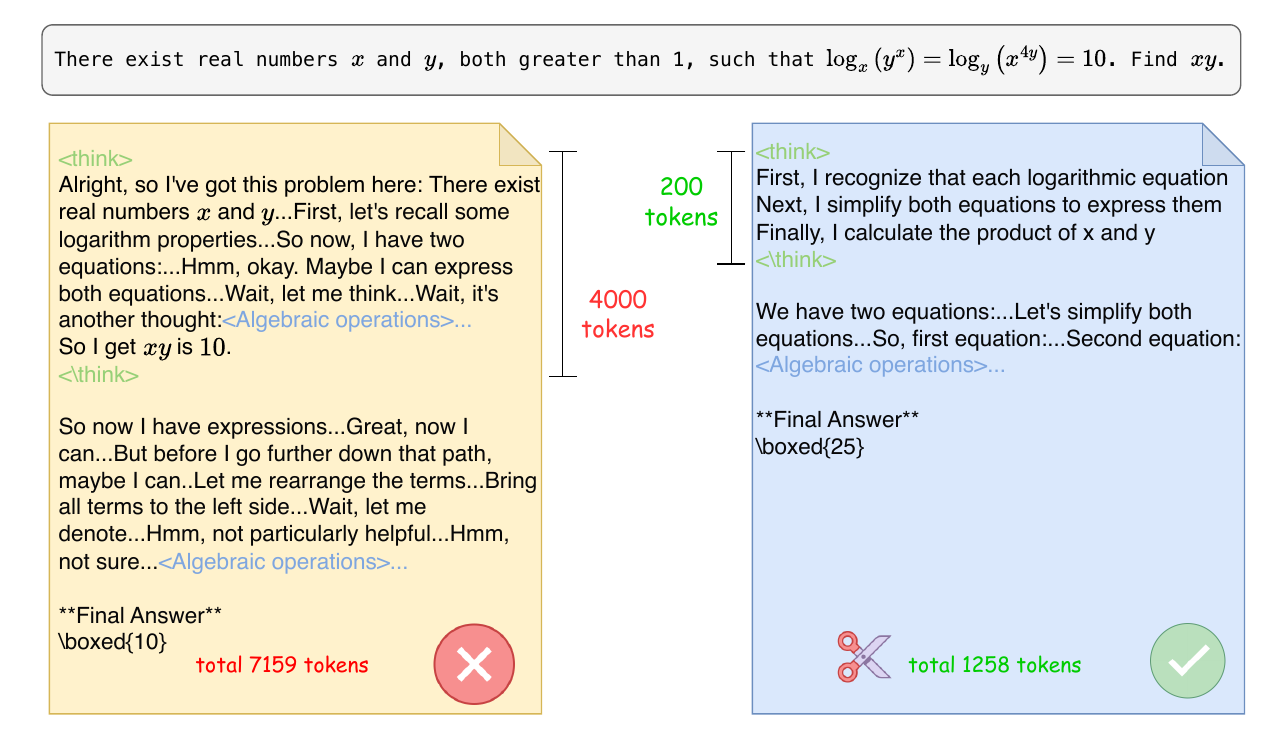}
    \caption{Case study on AIME24.}
    \label{fig:case_study}
\end{figure*}

\section{Discussion}
\label{apx:discussion}

Although our work focuses on the compression setting, the entropy-conflict analysis provides a more general perspective for understanding multi-objective optimization in large-scale model training. As discussed in §3.1, entropy conflict reflects a structural limitation: when a model is optimized toward a subset of objectives, it continues to preserve gradients associated with the original broader objective set, preventing full convergence to the subspace even when the new objective dominates.

This insight suggests broader implications for  RL-based LLM training pipelines. Many hybrid reinforcement learning objectives implicitly form nested objective sets—some goals are strict subsets of others. In such cases, jointly optimizing mixed objectives without considering this hierarchy may introduce gradient interference and slow specialization. A more efficient strategy is to explicitly exploit the inclusion structure: first train on the sub-objective set to let the model acquire the core behavioral rules, and then train on the superset to encourage free exploration under these learned constraints. Conceptually, the subset stage establishes the “rules of the game,” and the superset stage enlarges the space for exploration while ensuring the model remains anchored to the desired behaviors. This staged perspective offers a principled approach for improving controllability and sample efficiency in real-world model development.

\section{Case study}
\label{apx:case}

In this example(Figure \ref{fig:case_study}), the original model performs substantial computation inside the think phase: it repeatedly carries out algebraic manipulations, reevaluates earlier steps, and even restarts partial derivations before producing its final answer. After the think block ends, the model effectively recomputes the solution a second time in the visible output, resulting in severe redundancy and inflated reasoning length. In contrast, our compressed model adopts a fundamentally different behavior pattern. During the think phase it provides only high-level guidance—structuring the solution path without executing detailed calculations. All concrete computation is deferred to the final answer stage. This separation removes unnecessary internal loops, shortens the reasoning chain by eliminating duplicated work, and produces a more efficient and interpretable solution trajectory.